\documentclass[runningheads]{llncs}
\usepackage{graphicx}

\usepackage[ruled]{algorithm2e}
\usepackage{multirow}
\usepackage{booktabs}
\usepackage{graphicx}
\usepackage{orcidlink}

\usepackage{colortbl}
\usepackage{xcolor}

\begin{document}

\title{Particularity}

\author{Lee Spector\inst{1,2}\orcidlink{0000-0001-5299-4797} \and
Li Ding\inst{2} \and
Ryan Boldi\inst{2}}
\authorrunning{Spector et al.}

\institute{Amherst College, Amherst MA 01002, USA \and
University of Massachusetts, Amherst, Amherst MA 01002, USA
\email{lspector@amherst.edu}\\
}
\maketitle              
\begin{abstract}
We describe a design principle for adaptive systems under which adaptation is driven by particular challenges that the environment poses, as opposed to average or otherwise aggregated measures of performance over many challenges. We trace the development of this ``particularity'' approach from the use of lexicase selection in genetic programming to ``particularist'' approaches to other forms of machine learning and to the design of adaptive systems more generally.

\end{abstract}

\section{Overview}

In this paper we first describe lexicase selection, an algorithm  that was originally developed for use in genetic programming systems, which produce computer programs through processes of variation and selection.

We then present a generalization of the ideas that underlie lexicase selection, describing this generalization in terms of a design principle called ``particularity.'' 

After defining particularity, we review a sequence developments that exemplify particularity in different ways to extend the problem-solving power of the systems in which they are used. In doing so, we broaden our scope beyond genetic programming, discussing the use of particularity in other forms of machine learning, including deep neural networks, and in biology.

We conclude with some general comments about future of particularity in the design of adaptive systems.

\section{Lexicase}

In traditional approaches to genetic programming \cite{koza:book}, individuals are selected to serve as parents, and thereby to produce offspring, on the basis of scalar fitness values. Usually these fitness values are measures of performance over a collection of training examples, which are often called ``fitness cases'' in the genetic programming literature.

In some applications the different fitness cases that define a problem may present similar challenges to one another, and success or failure on any one case may be roughly as informative as success or failure on any other. But in many applications this will not be true. In some, edge cases of several kinds may be present, and some cases may require qualitatively different approaches than others. The numbers of cases that call for each approach may vary, and the number and nature of categories of cases may be unknown.

Lexicase selection \cite{Spector:2012:GECCOcompA,Helmuth:2015:ieeeTEC} is a parent selection method that prioritizes a single fitness case first and foremost when selecting each parent. It breaks ties with a second single fitness case, and then a third, and so on until a winner emerges.\footnote{This {\bf lexi}cographic processing of fitness {\bf case}s is the reason that lexicase selection is so named.} If several potential parents remain after considering all cases then the final tie is broken randomly. In the standard version of the technique the sequence of fitness cases used is random, with a different shuffle of the  cases used for selecting each parent.

Lexicase selection has been shown to significantly improve the problem-solving power of genetic programming in settings ranging from digital circuit design to general software synthesis, and in some settings it has been shown to allow genetic programming systems to solve problems that could not be solved when using selection methods based on scalar fitness measures  \cite{Helmuth:2013:GECCOcomp,Helmuth:thesis,Helmuth:2015:ieeeTEC,Helmuth:2015:GECCO,Helmuth:2021:GECCO,Tetteh:2021:EuroGP}.

Over evolutionary time, lexicase selection focuses on each particular case, each pair of cases, and more generally each subset of a problem's fitness cases.  It focuses on each in the sense that eliteness with respect to each will, with some shuffle of the cases, allow an individual to produce offspring. This is because each subset of the fitness cases may sometimes occur before all other cases in the shuffle, and thereby determine which individuals can be selected as parents.

Note that lexicase selection behaves differently than methods that select on the basis of generally good performance over many cases. Lexicase selection will, for example, often select individuals that are ``specialists'' in the sense that they are elite on one or a small number of cases, but atrociously bad on many others. This promotion of specialists appears to be connected to the problem-solving benefits of lexicase selection \cite{Helmuth:GPEM:lexi}. 

Lexicase selection also behaves differently than methods that adjust the influence that each case can have on an individual's scalar fitness value, such as ``implicit fitness sharing'' \cite{McKay:2000:GECCO}. Prior work has demonstrated problem-solving advantages of lexicase selection over implicit fitness sharing \cite{Helmuth:2015:ieeeTEC}, with one explanation being that implicit fitness sharing cannot reward good performance on particular combinations of cases that are rarely handled well by the same individual. 

Prior work has also considered the co-solvability of pairs of cases \cite{Krawiec:ppsn2010}, along with mechanisms designed to maintain diversity with respect to user-specified qualities \cite{mouret_illuminating_2015,pugh_quality_2016}. These techniques can be considered to embrace the particularity design principle to some extent. While detailed comparisons of these techniques to lexicase selection are beyond the scope of this paper, we note that lexicase selection is perhaps both simpler to implement and more thoroughgoing in its particularity.

We define the ``particularity" design principle as a mandate to prefer design choices, like those embodied in lexicase selection, that take all particular challenges and combinations of challenges posed by the environment seriously, and to explore their implications as independently as possible while avoiding averaging over multiple challenges. 

Often, in the history of the design of adaptive systems, averaging or other forms of aggregation have been considered necessary because the widely-used optimization methods operated only with scalar objective functions. But we now know that in many contexts there are alternatives, for example with lexicase selection or other many-objective optimization techniques. By attending more closely to the particular challenges posed by the environment, these methods may be able to adapt more quickly and successfully.

\section{Variance}

The particularity of lexicase selection, meaning its promotion of good performance on each particular environmental challenge and on each particular combination of environmental challenges, contrasts not only with the common practice in genetic programming but also with a wide range of practices in machine learning more generally, where averaging and other forms of aggregation are ubiquitous.

Many machine learning methods do allow users to specify hyperparameters that adjust the ``bias-variance tradeoff,'' meaning the extent to which individual training examples influence a model's behavior. 

High variance configurations allow each training example to have a large influence, but in many machine learning settings that means that the influence is on the behavior of a single model that is being trained. In evolutionary algorithms that operate on populations, by contrast, the influence can be on one among many approaches toward solutions that are being explored simultaneously in the population. 

In a sense, particularist methods, when used in population-based algorithms, allow us to manage the bias/variance trade-off by refusing to sacrifice one for the other, demanding instead that we prioritize both the guidance provided by individual challenges (high variance) and the guidance provided by larger collections of challenges (high bias) in different parts of the population. 

We note that higher population diversities often result from particularist methods, since different sub-populations are advantaged by attention to different combinations of features of the environment \cite{Helmuth:2015:GPTP}. Prior studies suggest, however, that the advantages of these methods stem not from the maintenance of diversity per se, but from the fact that the contours of the diverse populations they produce reflect the diversity of the challenges posed by the environment \cite{Helmuth:2016:GECCOcomp}. That is, the advantages stem from particularity. The extent to which such advantages can be obtained in settings that don't involve populations is a topic that deserves further study.

\section{Epsilon}

For some kinds of applications, the particularity of the original form of lexicase selection is too extreme. 

Consider, for example, continuous-valued symbolic regression applications. In this setting, the error for each individual on each case will be a real number, and it may be likely, depending on the problem and the genetic programming configuration, that no two individuals in the population will have exactly the same error for some particular case. 

When this happens there will be only one elite individual for the case in question, and there will never be ties that must be broken by performance on other cases. If this is true for a large number of cases then lexicase selection will usually select parents  on the basis of only a single case, or of a small number of cases. This means that there will be little focus on individuals that do well on combinations of cases, and little guidance of the evolutionary process toward individuals that can perform well on all cases.

Fortunately, epsilon lexicase selection solves this problem \cite{LaCava:2016:GECCO}. By allowing not only the elite to survive each filtering step of lexicase selection, but also individuals that under-perform the elite only by a small amount, epsilon lexicase selection can outperform scalar fitness-based methods in several settings \cite{DBLP:journals/corr/abs-1804-09331,LaCava:EC}.

Epsilon lexicase selection can be considered a ``relaxation'' of lexicase selection in the sense that the criteria for selection are less stringent. Other forms of relaxation have been explored, but not all of them appear to be advantageous \cite{Spector:2017:GPTP}. We may speculate that those that will be advantageous will be attentive to the particularity of the problem environment both with respect to individual cases and with respect to combinations of cases.

\section{Batched}

Another form of relaxation of lexicase selection involves the grouping of cases into batches, within which averaging or some other form of aggregation is performed. Parents are selected on the basis of shuffled sequences of batches rather than  shuffled sequences of single cases, with each step of filtering being based on comparisons of aggregate measures of performance over the batch.

Here the particularity of the method is applied not less stringently, as with epsilon lexicase selection, but rather more coarsely. This method proved successful in an application of lexicase selection in learning classifier systems, an evolutionary computing context quite different from genetic programming \cite{Aenugu:2019:GECCO}.

The success of batch lexicase selection in this setting suggests that intermediate levels of particularity may in some cases be helpful. It also suggests that the ``particulars'' upon which a particularist approach focuses need not be exactly the training examples provided by the problem environment. They might instead be ``cases'' that are derived from the training examples in some non-trivial way. Here they are simply averages over batches of training examples, but in principle, they might be derived from the training examples in other ways, some of which we discuss below.

\section{Downsampled}

What happens if we reduce the number of cases not by grouping them into batches over which we aggregate performance, but rather by using only the cases in a single batch, and considering them individually while ignoring all of the rest?

Downsampling is a general method by which data sets used for machine learning applications are decreased in size. Downsampling allows for larger systems to be used despite the entire data set being prohibitively expensive to enumerate. Furthermore, when downsampling is done every generation or iteration, it can help prevent overfitting as only portions of the training set are seen at a time, reducing the risk of memorization. 

Because downsampling constrains the set of challenges that can be seen by an individual at a given time, it can help an adaptive system to attend to the particularity of the sampled subset of its problem environment. For this to be successful in solving the environment's overarching problems, it is important that  samples are changed sufficiently often, and that the lessons learned from some samples can be maintained while exploring lessons learned from others.  

Another important effect of reducing the size of the training set is that every iteration becomes cheaper to perform. When downsampling to 10\% of the size of the training set, a similar number of iterations could be 10\% as expensive to perform. An evolutionary process could therefore be run for 10 times as long, or with 10 times as large a population, using the same computational budget.

When applied to lexicase selection, randomly downsampling training sets has been found to significantly improve problem-solving performance \cite{hernandez_random_2019,helmuth_problem_solving_2022,helmuth2022applying} when using the same computational budget as full lexicase selection. 

Selection schemes that select on the basis of aggregate measures seem to not benefit as much from the use of downsampling  as lexicase selection does \cite{boldi2023analyzing}. This may be because lexicase selection pays attention to particular challenges in the downsample, and is therefore able to maintain high levels of diversity that prevent premature convergence when running for a long time. 

One might think that changing the downsample entirely every generation, as it is done for random downsampling, could prevent effective learning. The intuition here is that the training set may be changed before the population really has a chance to get a foothold on the information that it provides. However, work has been done to show that, at least for genetic programming applied to certain benchmark program synthesis problems, the problem-solving power of these techniques is not hindered by rapid changes to the downsample \cite{boldi2022environmental}. The reason for this was attributed to the presence of synonymous or nearly synonymous cases that come in adjacent generations' downsamples. Synonymous cases are cases that measure similar behavior and as such are passed by similar groups of individuals. If each downsample is likely to contain cases that are synonymous with cases in the previous and next downsamples, then there will be some consistency in the challenges presented by the environment over evolutionary time.

\section{Informed}

One way of sharpening the focus of a downsampled selection scheme, as opposed to relaxing it, is to reduce the presence of synonymous cases. Synonymous cases are redundant as they share particularities with each other. Whilst including them might not be harmful on its own, they take the place of cases that might provide selection with more useful information.

\colorlet{lightgray}{gray!40}

\newcommand{\sig}[1]{\textbf{#1}}
\begin{table}[h!]
\centering
\setlength{\tabcolsep}{3.2pt}
\caption{Number of generalizing solutions (successes) out of 100 runs achieved by PushGP on the test set for a variety of program synthesis benchmark problems as reported in \cite{boldi_2023_informed}. Results are comparing the performance of Lexicase selection (Lex), Informed downsampled lexicase selection (IDS) and randomly downsampled lexicase selection (Rnd) on these problems. DS rate is the downsampling rate, or what proportion of the test cases appear in the sample each generation. The parent rate is the proportion of parents used to estimate the niche maintained by a sample, and the generational interval is the number of generations between we do this estimation. Problem names in \textbf{bold} face are where an informed downsampling approach performs the best out of all the techniques. Results signified with an asterisk (*) are significantly better than the corresponding run with random down-sampling at a $p{<}0.05$ level.
} 
\label{tab:push_results}
\renewcommand{\arraystretch}{1.3}
\begin{tabular}{l|c|ccccc|ccccc}
\hline
  \multicolumn{1}{l|}{\textbf{Method}} & \textbf{Lex} &  \textbf{Rnd}&  \multicolumn{4}{c|}{\textbf{IDS}}&  \textbf{Rnd} & \multicolumn{4}{c}{\textbf{IDS}} \\
  \multicolumn{1}{l|}{\textbf{DS Rate }} & \textbf{-} &   \multicolumn{5}{c|}{\textbf{0.05}} & \multicolumn{5}{c}{\textbf{0.1}} \\
  \multicolumn{1}{l|}{\textbf{Parent Rate}} & \textbf{-} &  \textbf{-}& \textbf{1} & \textbf{0.01} & \textbf{0.01} & \textbf{0.01} & \textbf{-}& \textbf{1} & \textbf{0.01} & \textbf{0.01} & \textbf{0.01}  \\
  \multicolumn{1}{l|}{\textbf{Gen. Interval}} & \textbf{-} &  \textbf{-}& \textbf{1} & \textbf{1} & \textbf{10} & \textbf{100} & \textbf{-}& \textbf{1} & \textbf{1} & \textbf{10} & \textbf{100}  \\
\hline
\hline

 \sig{Count Odds} &     24 &    25 &    43* &    99* &    \sig{100*} &    98* &    26 &   55* &    95* &    \sig{99*} &    97*  \\

\rowcolor{lightgray}
 \textbf{Find Pair} &    5 &    27 &    9 &    32 &    32 &    \textbf{36} &    15 &    7 &    19 &    19 &    \textbf{21}    \\

\sig{Fizz Buzz} &    13 &    64 &    2 &    85* &   94* &    \sig{95*} &    45 &    3 &    75 &    78 &    \sig{81*}    \\ 

\rowcolor{lightgray}
\sig{Fuel Cost} &     41 &    72 &    1 &    83 &   \textbf{85} &    83 &    \textbf{76} &    7 &    69 &    72 &    70    \\ 

 \sig{GCD}  &     20 &    74 &    4 &    \textbf{76} &    67 &    69 &    54 &    6 &    56 &    \textbf{63} &    62    \\

\rowcolor{lightgray}
 Grade &    0 &    0 &    0 &    0 & \sig{1} &    0 &    \sig{1} &    0 &  0 &    \sig{1} &    \sig{1}    \\

 \sig{Scrabble Score} &    8 &    8 &    6 &    69* &    64* &    \sig{75*} &    16 &    9 &    55* &    \textbf{74*} &    64*    \\

\rowcolor{lightgray}

\underline{\smash{Small or Large}} &     34 &    \textbf{93} &    37 &    69 &    69 &    69 &    \textbf{69} &    39 &    60 &    66 &    54   \\ 
\hline
\end{tabular}
\end{table}

Informed downsampled lexicase selection is a method to automatically detect and maintain a downsample of cases that are individually particular or unique \cite{boldi_2023_informed}. For example, the downsample would be filled with cases that measure qualitatively different behaviors in the individuals being selected. The downsamples are selected by analyzing how the population performs on training cases over the course of learning. If two training cases are passed by the same groups of individuals, then these test cases add no new information over each other. If two training cases are solved by disjoint sets of population members, these cases probably measure different behaviors, and are individually important to include in our training sets.

It turns out constructing downsamples using this information further improves the success rate of genetic programming runs that use downsampled lexicase selection \cite{boldi_2023_informed,boldi2023analyzing}. This benefit is likely due to maintaining higher test case coverage over the course of a run \cite{boldi_2023_static}. Because lexicase selection is able to focus on the particularities in a training set, having specific cases missing results in the loss of certain ecological niches. When all (or as many as possible) of the particularities are represented in the sample, lexicase selection can maintain the niches and more effectively pursue paths to solutions.

\section{Weighted}

Separately from the stringency, granularity, or downsampling of  lexicase selection's particularity, one can intervene in the order with which the evolutionary process is confronted with particular challenges.

In most work with lexicase selection, fitness cases have been chosen randomly, with each case having an equal chance of being chosen at each step in the selection process. However, recent work has shown that it can sometimes be helpful to provide more structure to this process, using weighted shuffles \cite{troise2018lexicase}. Rather than employing a uniformly random shuffle, weighted shuffling techniques skew the final arrangement of cases according to a specific metric. 

Additional work has demonstrated that weighted shuffle methods can enhance the efficiency of lexicase selection, using an approach called ``fast lexicase selection'' that integrates lexicase selection, weighted shuffles, and partial evaluation \cite{ding2022going,ding2022lexicase}. Experiments involving both genetic programming and deep learning tasks suggest that this method can significantly decrease the number of evaluation steps required to find solutions.

\section{Gradient}

\colorlet{lightgray}{gray!40}
\renewcommand{\arraystretch}{1.3}
\begin{table}[t]
    \caption{Comparing gradient lexicase selection to stochastic gradient descent (SGD) and other selection methods on CIFAR-10. We include the results reported in \cite{ding2022optimizing}, which shows that gradient lexicase selection can consistently improve the generalization performance of various popular network architectures.}
    \label{tab:gradlexi}
    \centering
    \setlength{\tabcolsep}{7pt}
    \begin{tabular}{lcccccccc}
    \toprule
    \multirow{2}{*}{Architecture} & \multicolumn{2}{c}{SGD} & \multicolumn{2}{c}{Random} & \multicolumn{2}{c}{Tournament} & \multicolumn{2}{c}{Lexicase} \\
    \cmidrule(r){2-3}
    \cmidrule(r){4-5}
    \cmidrule(r){6-7}
    \cmidrule(r){8-9}
    & \textit{acc.} & \textit{std.} & \textit{acc.} & \textit{std.} & \textit{acc.} & \textit{std.}  & \textit{acc.} & \textit{std.} \\
    \midrule
    VGG16 & 92.85 & 0.10 & 92.97 & 0.15 & 93.12 & 0.12 & \textbf{93.40} & 0.13 \\
    \rowcolor{lightgray} ResNet18 & 94.82 & 0.10 & 94.99 & 0.12 & 94.90 & 0.14 & \textbf{95.35} & 0.06 \\
    ResNet50 & 94.63 & 0.46 & 94.75 & 0.13 & 94.77 & 0.04 & \textbf{94.98} & 0.18 \\
    \rowcolor{lightgray} DenseNet121 & 95.06 & 0.31 & 95.13 & 0.04 & 95.12 & 0.02 & \textbf{95.38} & 0.04 \\
    MobileNetV2 & \textbf{94.37} & 0.19 & 94.02 & 0.14 & 93.91 & 0.09 & 93.97 & 0.12 \\
    \rowcolor{lightgray} SENet18 & 94.69 & 0.14 & 95.04 & 0.15 & 95.01 & 0.23 & \textbf{95.37} & 0.23 \\
    EfficientNetB0 & 92.60 & 0.18 & 92.77 & 0.11 & 92.83 & 0.12 & \textbf{93.00} & 0.22\\
    \bottomrule
    \end{tabular}
\end{table}

A variety of efforts have been taken to apply lexicase selection to adaptive systems outside of genetic programming, for example with the applications to learning classifier systems described above, and applications to fixed-length genetic algorithms for solving Boolean constraint satisfaction problems~\cite{metevier:2018:GPTP}. Recently, it has also been applied to the training of deep neural networks, where it has been shown to have utility for improving generalization \cite{ding2022optimizing}. 

The technique of ``gradient lexicase selection'' combines gradient descent and lexicase selection by performing gradient descent on several copies of a model, using different subsets of the training examples for each copy, and then using lexicase selection to determine which of the resulting, partially trained models will serve as the parent for the next cycle of gradient descent and selection. This method improves the generalization of popular deep neural network architectures on image classification benchmarks (as shown in Table~\ref{tab:gradlexi}), and a qualitative analysis indicates that the method causes networks to learn more diverse representations.

Models that are selected to serve as parents in gradient lexicase selection may not have the best aggregated loss among their siblings, but they will excel with respect to specific combinations of challenges posed by their environments. As a result, we may expect them to learn feature representations that allow them to make highly accurate predictions in particular circumstances. Over the course of learning, all challenges and combinations will have a chance to drive this process which, we may hypothesize, may be better able to avoid local minima than search processes driven only by aggregated loss.

Gradient lexicase selection can be viewed as an adaptation of lexicase selection to modern optimization tasks where gradient information is essential for learning and the dataset is large-scale. Work to date demonstrates the advantage of a particularity design in gradient-based optimization, and the results suggest that it will be useful to extend the idea to other forms of machine learning.

\section{Plexicase}

Theoretical analysis of the lexicase and epsilon lexicase selection methods has cast light on the expected probabilities of selection under these methods \cite{LaCava:EC}, although other work has shown that calculation of the exact probability distributions is NP-hard \cite{dolson-NP-Hard}. Specifically, the recurrence in lexicase selection events ultimately generates a probability distribution indicating which individuals will likely be chosen. However, this recursive process creates incremental dependencies, making it challenging to directly compute the probability distribution associated with the selection of individuals.

\begin{figure}[t]
    \begin{center}
      \includegraphics[width=.8\linewidth]{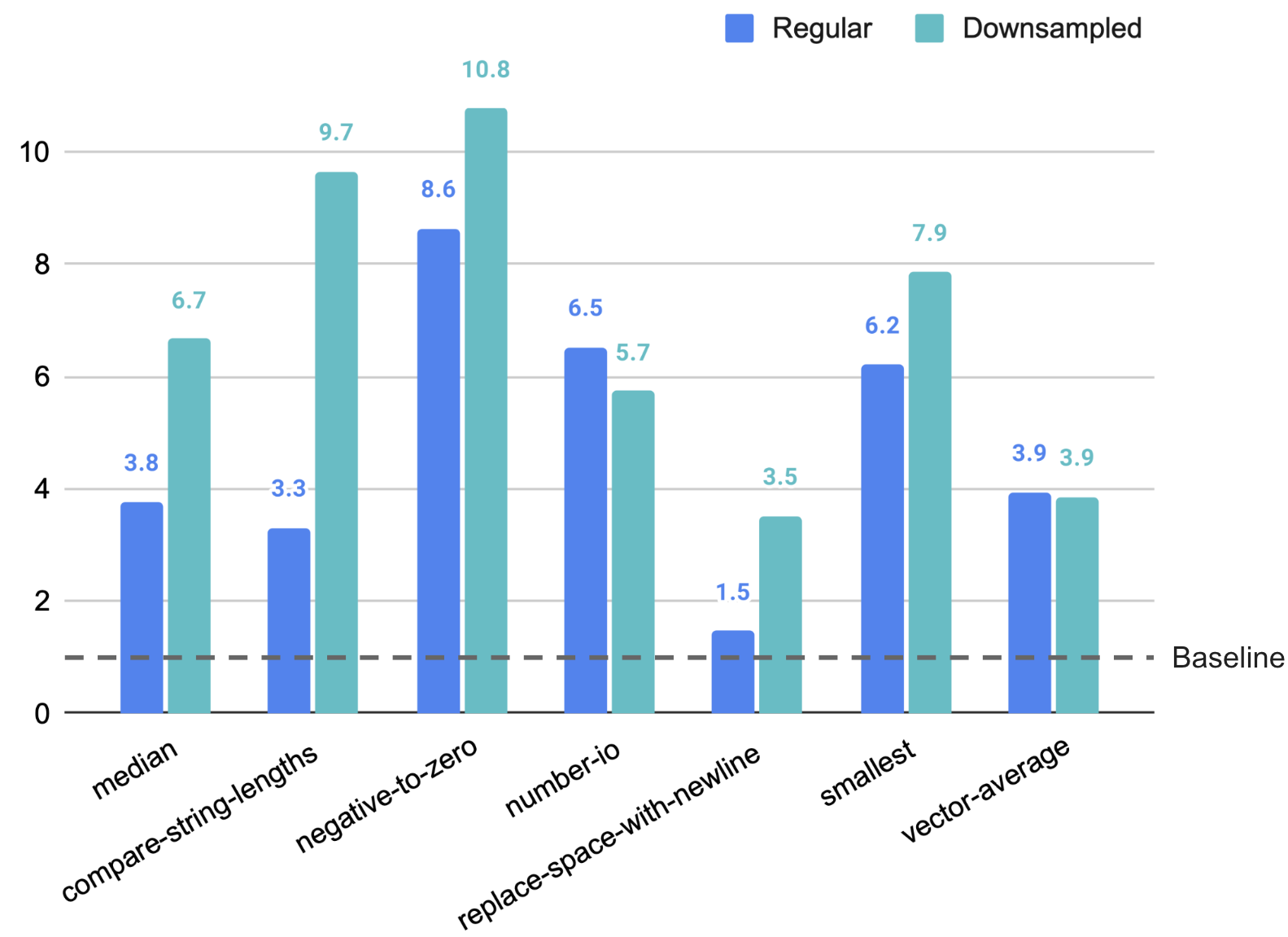}
    \end{center}
    \caption{Average runtime speed-up of plexicase selection compared to lexicase selection on GP problems in PSB~\cite{helmuth2015general}. A speed-up of $n$ means the plexicase selection runtime is $1/n$ of the lexicase selection runtime. The comparisons are conducted in both regular and downsampled circumstances. Figure adapted from \cite{ding:2023:GECCO}.
    }
    \label{fig:gptime}
\end{figure}

Building on this work, an approximation method has been developed that, while not exactly duplicating the selection probabilities of lexicase selection, can be computed much more quickly. On widely-used program synthesis and symbolic regression problems, this probabilistic lexicase selection method, called plexicase selection,
performs nearly as well lexicase selection in terms of the problems it can solve while significantly outperforming lexicase selection in terms of efficiency \cite{ding:2023:GECCO}. This technique efficiently determines an approximation of the selection probability distribution for individuals under lexicase selection, and it samples individuals from this distribution as an alternative to the filtering  process normally used for lexicase selection.

Plexicase selection holds two primary advantages. Firstly, it calculates the individuals' selection probability directly, rather than through repeated selection events, which can notably decrease the algorithm's runtime. Secondly, possessing the individuals' probability distribution facilitates parametric optimization in the selection process, which can be used to tune the selection pressure to further improve the algorithm performance. Experimental results in two domains (program synthesis and symbolic regression) demonstrated that plexicase selection significantly outperforms lexicase selection in terms of efficiency, and at the same time demonstrates superior or competitive problem-solving capabilities.

In general, these studies show that the specific methods used in lexicase selection, of lexicographic filtering based on randomly ordered training cases, are not the only way to achieve the adaptive particularity of lexicase selection.

\section{Hidden}

\begin{figure}[t]
\centering
\tikzset{every picture/.style={line width=0.75pt}} 

\begin{tikzpicture}[x=0.75pt,y=0.75pt,yscale=-1,xscale=1]

\draw   (53,21.5) -- (146,49.4) -- (146,105.6) -- (53,133.5) -- cycle ;
\draw   (149.67,50.32) -- (167.33,50.32) -- (167.33,104.68) -- (149.67,104.68) -- cycle ;
\draw   (265,133.5) -- (172,105.6) -- (172,49.4) -- (265,21.5) -- cycle ;

\draw (71,70) node [anchor=north west][inner sep=0.75pt]   [align=left] {Encoder};
\draw (25,70) node [anchor=north west][inner sep=0.75pt]    {$b( \theta )$};
\draw (192, 70) node [anchor=north west][inner sep=0.75pt]   [align=left] {Decoder};
\draw (268,70) node [anchor=north west][inner sep=0.75pt]    {$\hat{b}( \theta )$};
\draw (147,105) node [anchor=north west][inner sep=0.75pt]    {$ \begin{array}{l}
l( \theta )\\
\end{array}$};

\end{tikzpicture}
\caption{Variational auto-encoder architecture used to learn a latent encoding $l(\theta)$ of an individual's behavior in an environment $b(\theta)$. As this is a variational auto-encoder, $l(\theta)$ is sampled from a distribution that is parameterized by the output of the encoder (not pictured here for simplicity). Figure adapted from \cite{boldi2023QDBench}}\label{fig:vae1}
\end{figure}
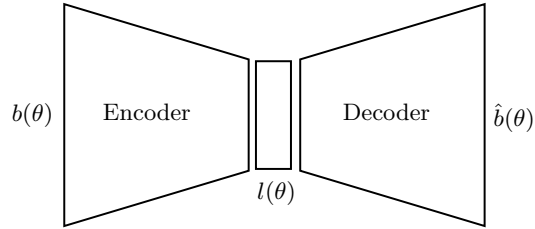

\begin{figure}[t]
\centering

\tikzset{every picture/.style={line width=0.75pt}} 

\begin{tikzpicture}[x=0.75pt,y=0.75pt,yscale=-1,xscale=1]

\draw   (80,40.9) -- (173,68.8) -- (173,125) -- (80,152.9) -- cycle ;
\draw   (175.67,69.72) -- (193.33,69.72) -- (193.33,124.08) -- (175.67,124.08) -- cycle ;

\draw (98,90) node [anchor=north west][inner sep=0.75pt]   [align=left] {Encoder};
\draw (50,90) node [anchor=north west][inner sep=0.75pt]    {$b( \theta )$};
\draw (172,125) node [anchor=north west][inner sep=0.75pt]    {$ \begin{array}{l}
l( \theta )\\
\end{array}$};
\draw (194,90) node [anchor=north west][inner sep=0.75pt]    {$ \begin{array}{l}
p( \theta )\\
\end{array}$};

\end{tikzpicture}
\caption{Using the previously trained encoder to predict the particularities $p(\theta)$ of an individual acting in an environment with behavior $b(\theta)$. These particularities are either directly equal to the latent layer $l(\theta)$ or some function of that layer. Figure adapted from \cite{boldi2023QDBench}}\label{fig:vae-measures}
\end{figure}
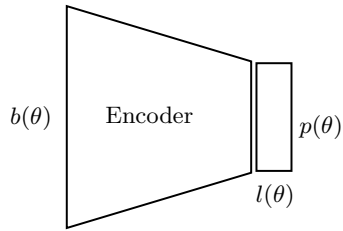

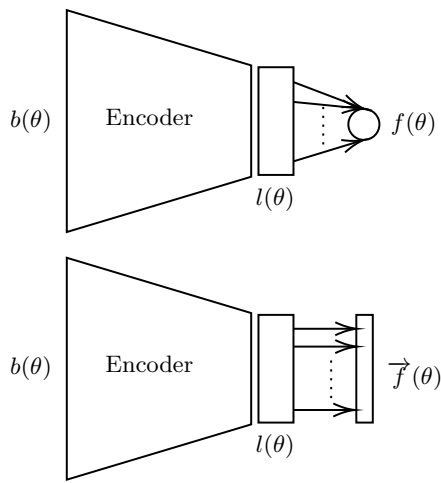
\begin{figure}
\centering

\tikzset{every picture/.style={line width=0.75pt}} 

\begin{tikzpicture}[x=0.75pt,y=0.75pt,yscale=-1,xscale=1]

\draw   (100,10) -- (193,37.9) -- (193,94.1) -- (100,122) -- cycle ;
\draw   (196.67,38.82) -- (214.33,38.82) -- (214.33,93.18) -- (196.67,93.18) -- cycle ;
\draw   (242,67.67) .. controls (242,63.34) and (245.51,59.83) .. (249.83,59.83) .. controls (254.16,59.83) and (257.67,63.34) .. (257.67,67.67) .. controls (257.67,71.99) and (254.16,75.5) .. (249.83,75.5) .. controls (245.51,75.5) and (242,71.99) .. (242,67.67) -- cycle ;
\draw    (214.33,46.33) -- (247.96,59.12) ;
\draw [shift={(249.83,59.83)}, rotate = 200.82] [color={rgb, 255:red, 0; green, 0; blue, 0 }  ][line width=0.75]    (10.93,-3.29) .. controls (6.95,-1.4) and (3.31,-0.3) .. (0,0) .. controls (3.31,0.3) and (6.95,1.4) .. (10.93,3.29)   ;
\draw    (214.33,56.33) -- (247.84,59.64) ;
\draw [shift={(249.83,59.83)}, rotate = 185.63] [color={rgb, 255:red, 0; green, 0; blue, 0 }  ][line width=0.75]    (10.93,-3.29) .. controls (6.95,-1.4) and (3.31,-0.3) .. (0,0) .. controls (3.31,0.3) and (6.95,1.4) .. (10.93,3.29)   ;
\draw    (214.33,86.33) -- (247.92,76.08) ;
\draw [shift={(249.83,75.5)}, rotate = 163.03] [color={rgb, 255:red, 0; green, 0; blue, 0 }  ][line width=0.75]    (10.93,-3.29) .. controls (6.95,-1.4) and (3.31,-0.3) .. (0,0) .. controls (3.31,0.3) and (6.95,1.4) .. (10.93,3.29)   ;
\draw  [dash pattern={on 0.84pt off 2.51pt}]  (229.33,58.83) -- (229.33,79.83) ;
\draw   (100,135) -- (193,162.9) -- (193,219.1) -- (100,247) -- cycle ;
\draw   (196.67,163.82) -- (214.33,163.82) -- (214.33,218.18) -- (196.67,218.18) -- cycle ;
\draw    (214.33,170.83) -- (244.33,170.83) ;
\draw [shift={(246.33,170.83)}, rotate = 180] [color={rgb, 255:red, 0; green, 0; blue, 0 }  ][line width=0.75]    (10.93,-3.29) .. controls (6.95,-1.4) and (3.31,-0.3) .. (0,0) .. controls (3.31,0.3) and (6.95,1.4) .. (10.93,3.29)   ;
\draw    (214.33,179.83) -- (244.33,179.83) ;
\draw [shift={(246.33,179.83)}, rotate = 180] [color={rgb, 255:red, 0; green, 0; blue, 0 }  ][line width=0.75]    (10.93,-3.29) .. controls (6.95,-1.4) and (3.31,-0.3) .. (0,0) .. controls (3.31,0.3) and (6.95,1.4) .. (10.93,3.29)   ;
\draw    (214.33,211.83) -- (243.33,211.83) ;
\draw [shift={(245.33,211.83)}, rotate = 180] [color={rgb, 255:red, 0; green, 0; blue, 0 }  ][line width=0.75]    (10.93,-3.29) .. controls (6.95,-1.4) and (3.31,-0.3) .. (0,0) .. controls (3.31,0.3) and (6.95,1.4) .. (10.93,3.29)   ;
\draw  [dash pattern={on 0.84pt off 2.51pt}]  (233.33,186.83) -- (233.33,206.5) ;
\draw   (246.02,163.82) -- (254.33,163.82) -- (254.33,218.18) -- (246.02,218.18) -- cycle ;

\draw (118,183) node [anchor=north west][inner sep=0.75pt]   [align=left] {Encoder};
\draw (70,183.4) node [anchor=north west][inner sep=0.75pt]    {$b( \theta )$};
\draw (192,218.4) node [anchor=north west][inner sep=0.75pt]    {$ \begin{array}{l}
l( \theta )\\
\end{array}$};
\draw (260,185.9) node [anchor=north west][inner sep=0.75pt]    {$\overrightarrow{f}( \theta )$};
\draw (118,58) node [anchor=north west][inner sep=0.75pt]   [align=left] {Encoder};
\draw (70,58.4) node [anchor=north west][inner sep=0.75pt]    {$b( \theta )$};
\draw (192,93.4) node [anchor=north west][inner sep=0.75pt]    {$ \begin{array}{l}
l( \theta )\\
\end{array}$};
\draw (262,59.9) node [anchor=north west][inner sep=0.75pt]    {$f( \theta )$};

\end{tikzpicture}
\caption{Learning the weights for a linear combination of features that sums to an approximation for the true fitness function $f(\theta)$. These weights can then be used to predict fitness from particularities $p(\theta) = l(\theta)$. Then, we can de-aggregate the last layer of the learned fitness model to result in a set of sub-objectives $\overrightarrow{f}(\theta)$ that sum to an approximation of the ground truth fitness $f(\theta)$. Figure adapted from \cite{boldi2023QDBench}}\label{fig:fitness-model}
\end{figure}

As noted above, genetic programming problems are often expressed in terms of training examples known as fitness cases. Lexicase selection uses these cases as the basis for the particularity upon which selection is based. 

In other settings, however, we may not be given training examples, or the training examples that we are given may not provide the most helpful particularized basis for selection.

In many such settings we may be able to derive the particularities of the problem environment from the environment itself. This may allow us to apply lexicase selection, or other particularized methods such as plexicase selection, to problems that do not explicitly provide appropriate sets of training examples. It may also allow us to apply these methods more effectively to problems that {\it do} explicitly provide  training examples, if the derived particularities provide a better basis for selection than the provided examples. 

Consider attempting to apply the particularity principle to a reinforcement learning (RL) domain such as navigating a maze using a robot with a variety of sensors. Perhaps the only input received in this domain is the raw values of an agent's sensors. What particularities should we pay attention to? A single objective measure such as reward would not be particular enough, and perhaps taking the raw sensor information could be too high dimensional. In order to find a level of description that allows us to efficiently learn a high quality policy, we can learn a lower dimensional encoding from the raw sensor data. This lower dimensional encoding may give us more information regarding the specific behavior of the agent than pure reward, while not being intractably high dimensional.

A simple way to learn a lower dimensional encoding and thus extract the particularities from an environment is to use an auto-encoder \cite{boldi2023QDBench}. Figure~\ref{fig:vae1} outlines the base auto-encoder architecture. Given a phenotype (or agent-environment sensory data), we can learn to reconstruct the same phenotype after compressing it through a bottleneck layer. Once the auto-encoder is trained, the features that are produced as output from the encoder are a lower dimensional representation of the agent's behavior. Using the encoder of this trained model to predict features (that could be used as measures for quality diversity optimization) \cite{mouret_illuminating_2015,pugh_quality_2016}\ is outlined in figure~\ref{fig:vae-measures}. One issue here is that the features that we extract might not be qualities that we want to promote (i.e. correlated with fitness). We can augment the model to output a set of features that are positively correlated with fitness. To do this, the encoder is frozen, and one (or perhaps more) layer can be added and trained to predict fitness from the output of the encoder (figure~\ref{fig:fitness-model}). This single node in the final layer will then be computing a sum of weighted features, where each feature's weight is its contribution to fitness. Then, disaggregating this sum (i.e. changing the dot product to an element wise product) allows us to receive a vector of features that are each importantly correlated with the fitness of the agent. This model allows for efficient extraction and weighting of particularities directly from the environment.

Returning to the RL example given above, we can start to see how such an extraction of particularities could be helpful for a learning agent. Perhaps some of the features extracted by the encoder are ``distance to closest wall," ``orientation," ``speed" or other high level features. Then, weights are learned to make each of these features correlated with fitness. For the feature of ``distance to closest wall" the weight would probably be negative, as getting close walls likely leads to the low fitness event of a crash. ``Speed" would be an example of a feature with a positive weight. The positively weighted features (or negated versions of the negative weighted features) are things we want to maximize in order to perform well in the domain.

The extraction of particularities does not need to involve an auto-encoder. We simply require a system that can efficiently extract a representation of the challenges posed by an environment. Whilst running the environment through an auto-encoder and determining which of its features correlate with challenges (fitness) is potentially effective, it is not the only way to achieve the same goal. Consider a neural network reward function, where the input is the behavior or phenotype of an individual, and the output is a predicted reward value. If this network works, simply de-aggregating the penultimate layer would create a series of fitness features that contribute to the overall fitness. Importantly, summing these de-aggregated fitness features together gives the predicted total fitness. 

\begin{figure}[t]
    \centering
    \includegraphics[width=\textwidth]{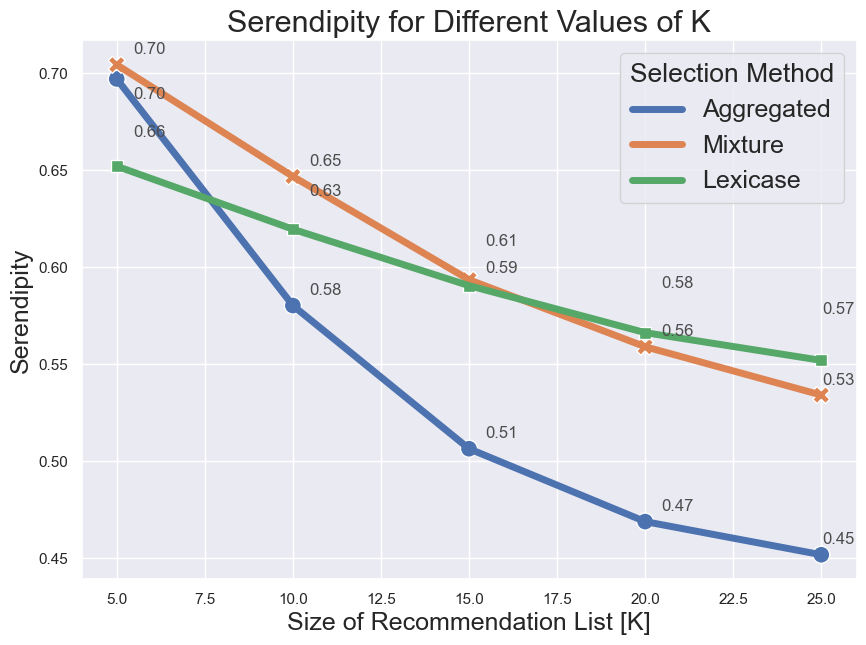}
    \caption{The serendipity (high novelty and match) of recommendations created by selecting items after neural matrix factorization with lexicase selection on the disaggregated preference features, selecting items with the highest aggregated scores, or a mixture of both (50/50). Figure adapted from \cite{boldi2023recsys}.}
    \label{fig:recsysresults}
\end{figure}

A similar technique has been used to improve serendipity of recommendations provided by recommendation systems to help prevent echo chambers in these systems. Serendipity is a metric that tracks whether items are unique but still high quality for a user (also known as surprise). This was done by disaggregating the penultimate layer of a neural network predicting how a user would rate an item. Then, items were lexicase selected based on this penultimate layer \cite{boldi2023recsys}. This improved the personalization, diversity and coverage of a series of recommendation lists in comparison to those produced using the aggregated predicted rating. Results for serendipity have been adapted from \cite{boldi2023recsys} and displayed in figure~\ref{fig:recsysresults}. These results highlight that lexicase maintains higher levels of serendipity for large recommendation list sizes. 

Using lexicase selection, or other methods that attend to particularities, could reduce the presence of echo chambers in networks that rely on recommendation systems. Particularized recommendation systems will recommend items that each maximally satisfy some combination of the user's preferences, and the recommendations will be diverse overall because different combinations of preferences will guide each recommendation. We see the potential here for systems that maximize the value of recommendation systems for both the platform and the users, while also creating richer, healthier, and more resilient information ecosystems.

\section{Living}

Selection in biology conforms to the particularity design principle. 

Organisms reproduce if they overcome the challenges that they face in the world. These challenges range from finding food and avoiding predators to completing all of the behavioral and molecular steps necessary for the production of a healthy child. 

Nowhere in this process does an individual organism's reproductive success depend on its average performance over a range of hypothetical challenges that it doesn't actually face in its lifetime. Each organism must contend with only the particular challenges that it encounters. A successful organism might be abysmally bad at dealing with challenges that it is fortunate enough not to face.

In addition, biological selection always takes place in the context of a population. Across the population, individuals generally confront somewhat different challenges. The particularity of biological selection means that reproductive success accrues to individuals that excel with respect to different combinations of challenges, and strategies rooted in a good performance on different combinations of challenges will be explored within the population simultaneously.

Does this mean that biological selection will always favor specialists over mediocre generalists, as lexicase selection does in genetic programming \cite{Helmuth:GPEM:lexi}? It does not appear to do so, and there is a large and rich literature exploring the evolution of specialists vs. generalists in biological systems. The extent to which generalists may be favored has been shown to depend on many factors including environmental heterogeneity \cite{https://doi.org/10.1046/j.1420-9101.2002.00377.x}, the relative speeds of reproduction and environmental change \cite{Sachdeva2020tuning}, and the phenotypic plasticity of individuals \cite{Fraebel2020plasticity}. This is still an area of active research. Conceivably, particularity-based analyses or simulations may contribute to work in this area.

In any case, biology's highly particularized search strategy has produced the most adaptive systems of which we have any knowledge. It seems therefore to be a good model on which to build new adaptive systems.

\section{Honor}

\begin{center}
Specificity is the soul of narrative.

---John Hodgman
\end{center}

One way to think about the particularity design principle is that it counsels us to design adaptive systems in ways that make them sensitive to the specific elements of the circumstances that they encounter, and the specific results of their actions in those specific circumstances. All of this is opposed to using averages or other forms of aggregation along the pathway from the environment back to the mechanisms of adaptation. As we have seen above, it may not always pay to take this advice too strictly. But the principle nonetheless suggests a preference for specificity.

Alternatively, and perhaps more memorably, the particularity design principle might be glossed as a mandate to ``honor all the things, and all the combinations of all the things.''

What are the things? Sometimes they are given in the form of training examples. Sometimes they may be batches of examples. Sometimes they may be learned features of the problem environment.

What does it mean to honor them? In lexicase selection it means to treat each of them as having primary importance  sometimes, and secondary importance sometimes, and so on. By doing this, via filtering  based on random sequences of fitness cases, lexicase selection maintains a population in which all of the things, and all of the combinations of the things, can serve as the foundation for the evolution of solutions. 

More generally, ``honoring'' here means to explore approaches to solutions that begin with a focus on each environmental feature, and on each combination of environmental features.

We have seen that such focus can enhance our ability to solve difficult problems in several settings. So far, however, we have only scratched the surface of exploring this approach. Within genetic programming there is clear evidence that it can be useful in a variety of contexts. Within the study of machine learning methods and adaptive systems more generally, there is much more work to be done to study this approach's effectiveness and range of applicability.

\section*{Acknowledgments}

We thank Bill Tozier, Anil Saini, Eddie Pantridge, Andrew Ni, Nic McPheee, Tom Helmuth, Ramita Dhamrongsirivadh, and other members of the Amherst College PUSH lab for stimulating conversations that helped us to develop the ideas described in this paper. We also thank participants in the 2023 Genetic Programming Theory and Practice workshop, and particularly Alex Lalejini, Erik Hemberg, and Joel Lehman, who commented on a draft. This material is based upon work supported by the National Science Foundation under Grant No. 2117377. Any opinions, findings, and conclusions or recommendations expressed in this publication are those of the authors and do not necessarily reflect the views of the National Science Foundation.
This work was also performed in part using high-performance computing equipment obtained under a grant from the Collaborative R\&D Fund managed by the Massachusetts Technology Collaborative.

\bibliographystyle{splncs04}
\bibliography{GPTP-2023-Spector}

\end{document}